\documentclass[10pt,twocolumn,letterpaper]{article}

\usepackage[pagenumbers]{cvpr} 
\usepackage{multirow}
\usepackage{makecell}
\usepackage{amssymb}
\usepackage[misc]{ifsym}
\pdfoutput=1
\usepackage[dvipsnames]{xcolor}

\definecolor{cvprblue}{rgb}{0.21,0.49,0.74}
\usepackage[pagebackref,breaklinks,colorlinks,citecolor=cvprblue]{hyperref}

\def\paperID{7505} 
\def\confName{CVPR}
\def\confYear{2024}

\begin{document}

\title{DiffMorpher: Unleashing the Capability of Diffusion Models for \\ Image Morphing}

\author{Kaiwen Zhang\textsuperscript{\normalfont 1,2*}
\and
Yifan Zhou\textsuperscript{\normalfont 3}
\and
Xudong Xu\textsuperscript{\normalfont 2}
\and
Xingang Pan\textsuperscript{\normalfont 3 \Letter}
\and
Bo Dai\textsuperscript{\normalfont 2}
\\
\vspace{-0.4cm}
\and
\textsuperscript{\normalfont 1}{\normalfont Tsinghua University} \
\quad\textsuperscript{\normalfont 2}{\normalfont Shanghai AI Laboratory} \
\quad\textsuperscript{\normalfont 3}{\normalfont S-Lab, Nanyang Technological University}
\vspace{3pt} \\
{\normalsize Project page: \href{https://kevin-thu.github.io/DiffMorpher_page/}{https://kevin-thu.github.io/DiffMorpher\_page/} \vspace{-6pt}}
}

\twocolumn[{
	\renewcommand\twocolumn[1][]{#1}
	\maketitle
	\vspace{-7mm}
	\begin{center}
		\includegraphics[width=\textwidth]{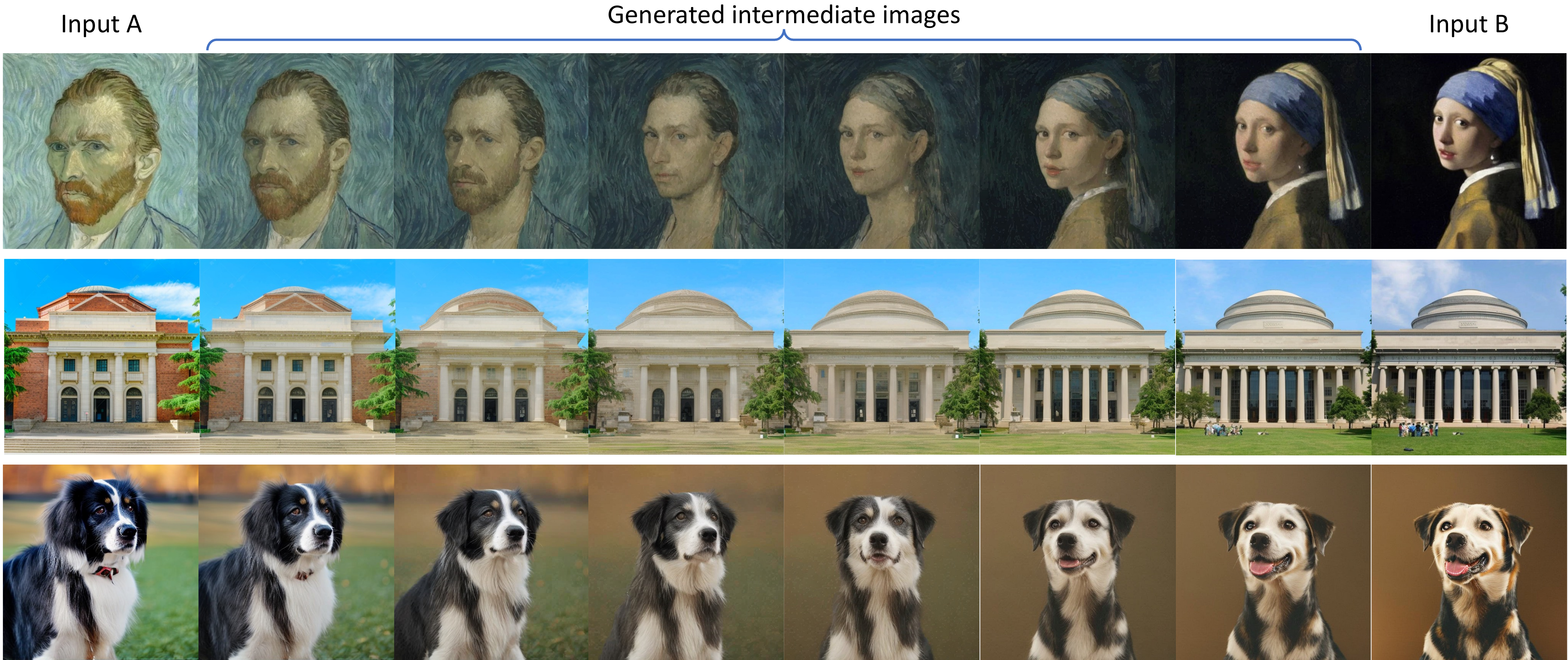}
	\end{center}
	\vspace{-6mm}
	\captionof{figure}{Given two input images, our approach can generate a sequence of intermediate images, delivering a smooth and natural transition between them. This is achieved purely by harnessing the prior knowledge of a pre-trained diffusion model, \ie, Stable Diffusion~\cite{LDM}.}
	\label{fig:teaser}
	\vspace{6mm}
}]

\maketitle


\begin{abstract}

\footnote{\textsuperscript{\normalfont *}Work done during internship at Shanghai AI Laboratory.}
\footnote{\textsuperscript{\normalfont \Letter} Corresponding Author.}

Diffusion models have achieved remarkable image generation quality surpassing previous generative models.
However, a notable limitation of diffusion models, in comparison to GANs, is their difficulty in smoothly interpolating between two image samples, due to their highly unstructured latent space.
Such a smooth interpolation is intriguing as it naturally serves as a solution for the image morphing task with many applications. 
In this work, we address this limitation via DiffMorpher, an approach that enables smooth and natural image interpolation by harnessing the prior knowledge of a pre-trained diffusion model.
Our key idea is to capture the semantics of the two images by fitting two LoRAs to them respectively, and interpolate between both the LoRA parameters and the latent noises to ensure a smooth semantic transition, where correspondence automatically emerges without the need for annotation.
In addition, we propose an attention interpolation and injection technique, an adaptive normalization adjustment method, and a new sampling schedule to further enhance the smoothness between consecutive images. 
Extensive experiments 
demonstrate that DiffMorpher achieves starkly better image morphing effects than previous methods across a variety of object categories, bridging a critical functional gap that distinguished diffusion models from GANs.

\end{abstract}

\section{Introduction}
Image morphing~\cite{Survey98, Survey17, Survey23} is a popular technique for image transformation, lying at the intersection of computer vision and computer graphics with continuous attention over decades.
Given two images of topologically similar objects and optionally a set of correspondence key points, a morphing process generates a sequence of reasonable intermediary images. When played in succession, the image sequence produces a captivating video of a smooth transition between the two input images.
Developed initially for cinematic and visual effects, image morphing has found its applications in various fields like animations, games~\cite{Survey17, Survey23}, as well as photo-editing tools~\cite{Survey23} for artistic and entertainment purpose to enrich people's imagination. 
In the era of deep learning, image morphing can also be used in data augmentation~\cite{MorphGAN}.

There are two major concerns in the problem of image morphing, which are hardly balanced in previous studies: the \textit{rationality} of intermediate images and the \textit{ smoothness} of the transition video. 
Classic methods presented in the graphics literature~\cite{FieldMorph,Triangle,Halfway,Patch01,Patch02} typically involve an image-warping process to align the correspondence points and a cross-dissolution operation to mix the colors.
However, the color-space dissolution does not well explain the textural and semantic transition and is prone to undesirable intermediate results like ghosting artifacts.
Since the deep learning era, GANs~\cite{goodfellow2014generative} have shown stunning image morphing ability through simple latent code interpolations~\cite{BigGAN, StyleGAN, StyleGAN2, StyleGAN3, StyleGANXL, StyleGANT, DGP}. 
Despite the smoothness of the transformation process, this method is hard to extend to arbitrary real-world images due to the limited model capacity and the challenge of GAN inversion~\cite{GANInversion}.
Recently, diffusion models~\cite{Diff15, DDPM, ScoreBased,LDM,Unidiffuser} have emerged as state-of-the-art generative models, significantly enhancing image synthesis and real image reconstruction. 
However, initial attempts to apply diffusion models to image morphing suffer from abrupt content changes between consecutive images. 

In this work, we are interested to address the image morphing problem by asking the question: \textit{Is it feasible to achieve smooth and natural image interpolation with diffusion models, akin to the capabilities of GANs?}
The solution to this problem will immediately serve as an image morphing approach when combined with image reconstruction techniques like DDIM inversion~\cite{DDIM}.
However, realizing such a reasonable interpolation on diffusion models is non-trivial.
Unlike GANs that have a meaningful compact latent space, the latent space of diffusion models is a noise map that lacks semantic meaning, thus random and abrupt content flickering are often observed when naively interpolating in the latent space.
How to guarantee smoothness in both high-level semantics and low-level textures remains a key challenge.

To this end, we present \textit{DiffMorpher}, a new approach to achieve \textit{smooth} image interpolation based on diffusion models while maintaining the \textit{rationality} of intermediate images.
Since the latent space is non-interpretable, our key idea is to create smooth semantic transition via the \textit{low-rank parameter space}.
This is achieved by applying low-rank adaptations (LoRAs)~\cite{Lora} to the two input images separately, encapsulating the corresponding image semantics in the two groups of LoRA parameters.
Thanks to the analogous parameter structures, a linear interpolation between the two sets of LoRA parameters will deliver a smooth transition in the image semantics.
Combining spherical interpolation (slerp) between the two latent Gaussian noises associated with the two input images, our approach can create a semantically meaningful transition with high-quality intermediates between them.
However, this method does not fully eliminate the low-level abrupt change.
To address this, we further introduce a self-attention interpolation and substitution method that ensures smooth transition in low-level textures, and an AdaIN adjustment technique that enhances the coherence in image colors and brightness.
Finally, to maintain a homogeneous transition speed in image semantics, we propose a new sampling schedule.

We extensively evaluate \textit{DiffMorpher} in a wide range of real-world scenarios.
A new image morphing benchmark \textit{MorphBench} is created to support quantitative evaluation, where our approach significantly outperforms existing methods in both smoothness and image fidelity.
To the best of our knowledge, this is the first time smooth image interpolation can be achieved on diffusion models at a comparable level as GANs.
Unlike GANs that struggle with real-world images, \textit{DiffMorpher} can deal with a much wider image scope.
The ability to continuously tweak image semantics has empowered GAN for many downstream applications, thus we hope our work will similarly pave the way for new opportunities in diffusion models.
For example, our method can augment many image editing methods such as~\cite{Imagic,DragonDiffusion,brooks2022instructpix2pix,zhang2023sine} by turning their final images into continuous animations.

\section{Related Work}

\subsection{Classic Image Morphing}

Image morphing is a long-standing problem in computer vision and graphics~\cite{Survey98, Survey17, Survey23}.
Classic graphical techniques~\cite{FieldMorph,Triangle,Halfway,Patch01,Patch02} typically combine correspondence-driven bidirectional image warping with blending operations to obtain plausible in-betweens in a smooth transition.
Although making a smooth morphing between two images, these methods fall short of creating new content beyond the given inputs, thus leading to unsatisfactory results like ghosting artifacts.
More recently, the explosion of data volume gave rise to a new data-driven morphing paradigm~\cite{DataDriven, MorphGAN}.
Unlike classic approaches, they capitalize on massive images from a specific object class to determine a smooth transition path from the source image to the target one, which contributes to compelling intermediate morphing results.
However, the great demand for enormous single-class data impedes their applications in more general scenarios like cross-domain or personalized morphing.
In contrast, our model leverages the prior knowledge in diffusion models pre-trained on large-scale images and thus is applicable to diverse object categories.

\subsection{Image Editing via Diffusion Models}
Diffusion models~\cite{Diff15, DDPM, ScoreBased} have been a prevalent star in deep generative models in recent years, thanks to their impressive sample quality and scaling ability~\cite{DiffBeatGAN, CascadedDM}. 
By learning to gradually denoise from Gaussian noises with a noise prediction UNet~\cite{Ronneberger2015UNetCN}, diffusion models can generate high-quality clean data that fits well with real data distribution.
Diffusion models trained on large-scale text-image pairs~\cite{Schuhmann2022LAION5BAO}, such as Imagen~\cite{Imagen} and Stable Diffusion~\cite{LDM}, have gained unprecedented success in text-to-image generation.
Therefore, they are suitable as a powerful prior for multiple editing tasks, including text-guided~\cite{hertz2022prompt2prompt, brooks2022instructpix2pix, Imagic, Cao2023MasaCtrlTM, Parmar2023ZeroshotImageTrans, PlugAndPlay} and drag-guided~\cite{DragDiffusion, DragonDiffusion} image manipulation.
Most of these works directly generate the final edited image, while the generation of a continuous animation like image morphing is much less explored in the literature of image diffusion models.


\subsection{Deep Interpolation}
It has been widely demonstrated that Generative Adversarial Networks (GANs)~\cite{goodfellow2014generative} can be used to morph images by interpolating latent codes.
Due to their highly continuous and discriminative latent embedding space, a linear interpolation among two latent codes will exhibit impressive image morphing results, as demonstrated in a large body of GAN papers~\cite{BigGAN, StyleGAN, StyleGAN2, StyleGAN3, StyleGANXL, StyleGANT}.
However, to morph between real images, the corresponding latent codes, which are often outside of GAN's latent distribution, must be obtained with GAN inversion and tuning techniques~\cite{GANInversion, DGP, PTI}.
Typically, the latent codes obtained struggle to recover the original real images.
Although the generator can be tuned to reconstruct the images, the rationality of the intermediates and the correctness of correspondence cannot be guaranteed.

Recent advances in diffusion models~\cite{Diff15, DDPM, ScoreBased} also show the potential to generate reasonable intermediate images through latent noise interpolations and text embedding interpolations~\cite{Unidiffuser, wang2023interpolating}. 
However, due to the highly unstructured image distribution learned in diffusion models, the generated transition videos often contain abrupt changes and inconsistent semantic content, which are unacceptable in the image morphing task.
Preechakul \etal~\cite{preechakul2021diffusion} proposed a Diffusion Autoencoder architecture that enables more reasonable image interpolation than the vanilla diffusion models, but this approach cannot be directly applied to the widely used vanilla diffusion models like Stable Diffusion and abrupt changes still remain.
In our work, we demonstrate the ability of diffusion models to generate smooth and natural morphing sequences using only the prior knowledge in pretrained text-to-image models.

Recently, a concurrent work~\cite{Yang2023IMPUSIM} has also studied the application of diffusion models for the image morphing task.
Compared to their approach, our method incorporates delicately designed self-attention control and AdaIN adjustment, which greatly diminish abrupt changes in textures and improve consistency in colors.
Furthermore, our approach fits a single LoRA for each image and interpolates between the LoRA parameters during morphing, thus increasing the versatility and flexibility of our method, such as applying morphing among multiple images. 

\begin{figure*}[t]
	\centering
	\includegraphics[width=15cm]{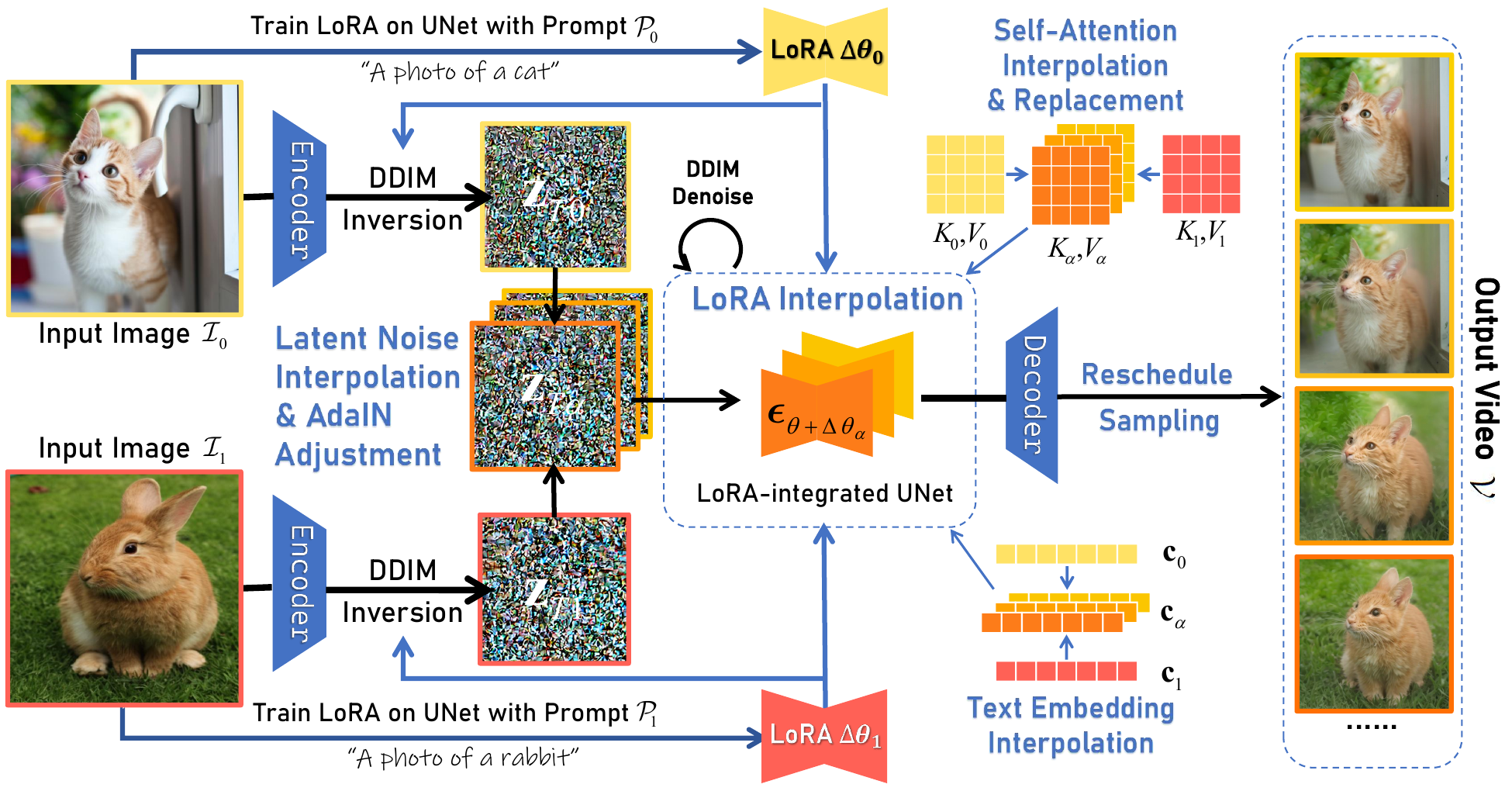}
        \vspace{-0.3cm}
	\caption{Method pipeline. Given two images $\mathcal{I}_0$ and $\mathcal{I}_1$, two LoRAs are trained to fit the two images respectively. Then the latent noises for the two images are obtained via DDIM inversion. The mean and standard deviation of the interpolated noises are adjusted through AdaIN. To generate an intermediate image, we interpolate between both the LoRA parameters and the latent noises via the interpolation ratio $\alpha$. In addition, the text embedding and the $K$ and $V$ in self-attention modules are also replaced with the interpolation between the corresponding components. 
 Using a sequence of $\alpha$ and a new sampling schedule, our method will produce a series of high-fidelity images depicting a smooth transition between $\mathcal{I}_0$ and $\mathcal{I}_1$.
	}
        \vspace{-0.3cm}
	\label{fig:method}
\end{figure*}

\section{Method}\label{Section:Method}
Given two images $\mathcal{I}_0$ and $\mathcal{I}_1$, our goal is to obtain an interpolation video $\mathcal{V}=\{\mathcal{I}_\alpha|\alpha \in (0,1)\}$ that displays a natural and smooth transition from $\mathcal{I}_0$ to $\mathcal{I}_1$, where the sequence of $\alpha$ depends on the desired number of frames $n$ and a specific sampling schedule.
A meaningful image morphing should be done between two images with clear correspondence.
In our general morphing framework, $\mathcal{I}_0$ and $\mathcal{I}_1$ can be either real images or diffusion-generated images with text prompts $\mathcal{P}_0$ and $\mathcal{P}_1$.

In this section, we formally present our \textit{DiffMorpher} approach to address this problem.
We first introduce the preliminaries on diffusion models in Sec. \ref{Sub:Prem}.
To capture the identities in $\mathcal{I}_0, \mathcal{I}_1$ and generate semantic consistent and meaningful in-betweens, we propose LoRA interpolation and latent noise interpolation techniques in Sec. \ref{Sub:LoRA} and \ref{Sub:Noise}.
To enhance the smoothness of the transition video, we propose the self-attention interpolation and replacement method, a AdaIN adjustment technique and a new reschedule method in Sec. \ref{Sub:Attn}, Sec. \ref{Sub:Adain} and \ref{Sub:Reschedule}.
An overview of our method with an illustration example is shown in Fig. \ref{fig:method}.

\subsection{Premininaries on Diffusion Models}\label{Sub:Prem}
Diffusion models~\cite{Diff15, DDPM, ScoreBased, DDIM} are a family of latent variable generation models of the form:
\begin{equation}
p_\theta(\mathbf{z}_0)= \int p_\theta(\mathbf{z}_{0:T})d\mathbf{z}_{1:T}
\end{equation}
It includes a diffusion process $\{q(\mathbf{z}_t) | t=0,1,\cdots, T\}$ that gradually adds noise to the data sampled from the real data distribution $q(\mathbf{z}_0)$ toward $q(\mathbf{z}_T)=\mathcal{N}(\mathbf{0},\mathbf{I})$, and a corresponding denoising process $\{p(\mathbf{z}_t)| t=T, T-1, \cdots, 0\}$ that generates clean data from the standard Gaussian noise $\mathbf{z}_T \sim p(\mathbf{z}_T)=\mathcal{N}(\mathbf{0},\mathbf{I})$, where $T$ is the total number of steps.
The denoising process is achieved by learning a parameterized joint distribution $p_\theta(\mathbf{z}_{0:T})$ with a noise prediction network $\mathbf{\epsilon}_\theta$. 
Specifically, in the denoising step $t$,  $\mathbf{\epsilon}_\theta$ predicts the noise $\epsilon$ added to $\mathbf{z}_{t-1}$ according to current noise $\mathbf{z}_t$, current time step $t$ and possible additional condition $\mathbf{c}$.
In practice, $\mathbf{\epsilon}_\theta$ is generally implemented as a UNet~\cite{Ronneberger2015UNetCN}.

Latent Diffusion Model (LDM)~\cite{LDM} is an important variant of diffusion models that achieves a great balance between image quality and sample efficiency.
Based on the LDM framework, a number of powerful pretrained text-to-image models have been available to the public, including the widely-used Stable Diffusion (SD).
It involves a variational auto-encoder (VAE)~\cite{VAE} that encodes the images to latent embeddings and trains a text-conditioned diffusion model in the latent space.
The denoising UNet $\mathbf{\epsilon}_\theta$ in the SD model is composed of a sequence of basic blocks, each of which includes a self-attention module, a cross-attention module~\cite{Vaswani2017AttentionIA}, and a residual block~\cite{ResNet}. 
The attention module in UNet can be formulated as follows:
\begin{equation}
Attention(Q, K, V) = softmax(\frac{QK^T}{\sqrt{d_k}})V
\end{equation}
where $Q$ is the query features derived from the spatial features, and $K, V$ are the key and value features obtained from either the spatial features (in self-attention layers) or the text embedding (in cross-attention layers) with respective projection matrices. 
Our method in this paper is built upon the SD model.

\subsection{LoRA Interpolation}\label{Sub:LoRA}
\begin{figure}[t]
        \includegraphics[width=8.5cm]{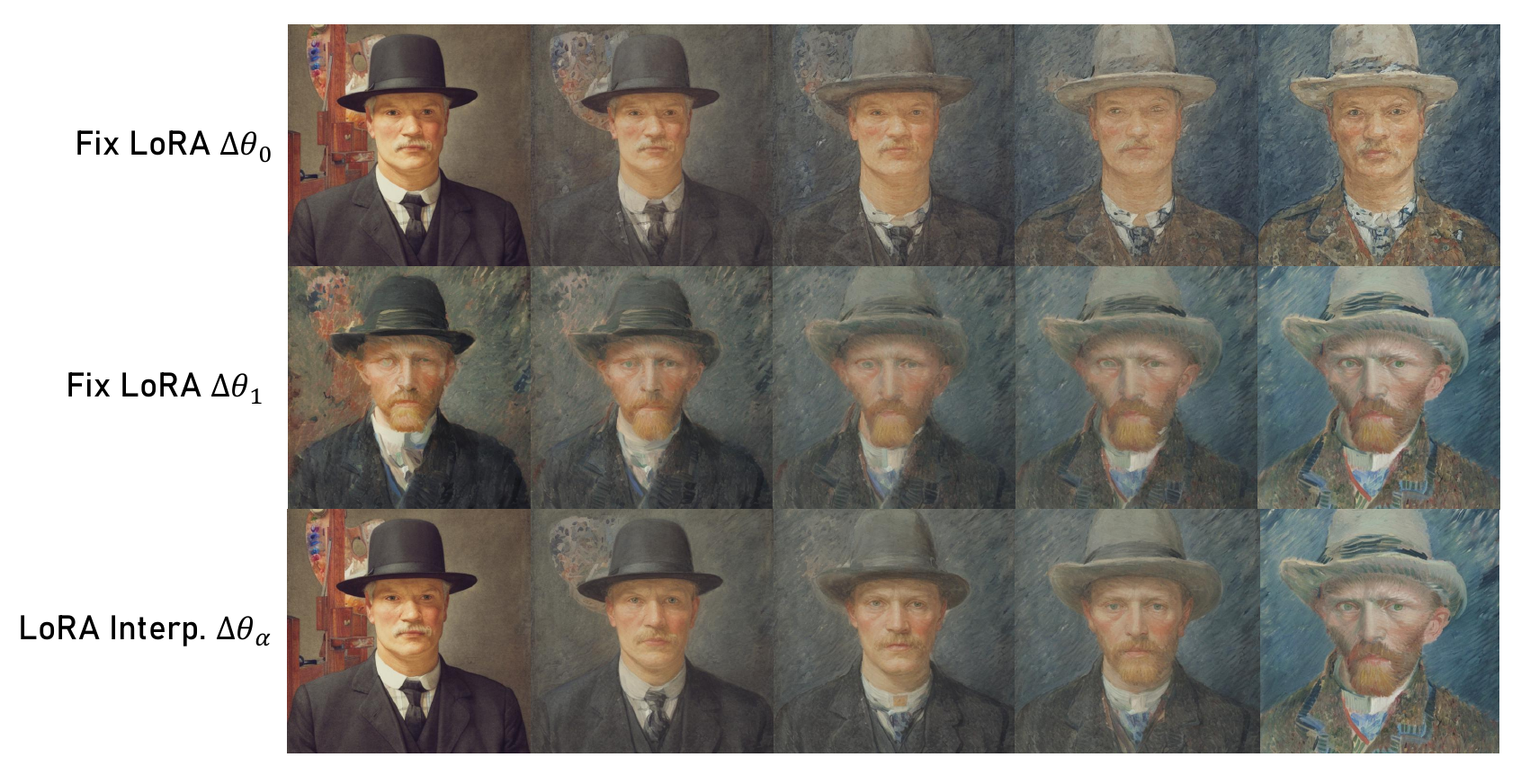}
        \vspace{-0.8cm}
	\caption{Effects of LoRA. A LoRA fit to an image tends to capture its semantic identity, while the layout and appearance are controlled by latent noise.}
        \vspace{-0.4cm}
	\label{fig:lora}
\end{figure}

Low-Rank Adaption (LoRA)~\cite{Lora} is an efficient tuning technique that was first proposed to fine-tune large language models and recently introduced to the domain of diffusion models. 
Instead of directly tuning the entire model, LoRA fine-tunes the model parameters $\theta$ by training a low-rank residual part $\Delta\theta$, where $\Delta\theta$ can be decomposed into low-rank matrices.
Besides its inherent advantage in training efficiency, we further discover that LoRA enjoys an impressive capacity to encapsulate high-level image semantics into the low-rank parameter space.
By simply fitting a LoRA on a single image, the fine-tuned model can generate diverse samples with consistent semantic identity when traversing the latent noise, as shown in Fig. \ref{fig:lora}.

Motivated by this observation, we first train two LoRAs $\Delta\theta_0$, $\Delta\theta_1$ on the SD UNet $\mathbf{\epsilon}_\theta$ for each of the two images $\mathcal{I}_0$ and $\mathcal{I}_1$. Formally, the learning objective for training $\Delta\theta_i (i=0,1)$ is:
\begin{equation}
\mathcal{L}(\Delta\theta_i)=\mathbb{E}_{\epsilon, t}[\Vert \epsilon-\mathbf{\epsilon}_{\theta+\Delta\theta_i}(\sqrt{\bar{\alpha}_t} \mathbf{z}_{0i}+\sqrt{1-\bar{\alpha}_t}\epsilon, t, \mathbf{c}_i)\Vert^2]
\end{equation}
where $\mathbf{z}_{0i}=\mathcal{E}(\mathcal{I}_i)$ is the VAE encoded latent embedding associated with the input image $\mathcal{I}_i$,
$\epsilon\sim\mathcal{N}(\mathbf{0},\mathbf{I})$ is the random sampled Gaussian noise,
$\mathbf{z}_{ti}=\sqrt{\bar{\alpha}_t} \mathbf{z}_{0i}+\sqrt{1-\bar{\alpha}_t}\epsilon$ is the noised latent embedding at diffusion step $t$,
$\mathbf{c}_i$ is the text embedding encoded from the text prompt $\mathcal{P}_i$,
and $\epsilon_{\theta+\Delta\theta_i}$ represents the LoRA-integrated UNet.
The fine-tuning objective is optimized separately via gradient descent in $\Delta\theta_0$ and $\Delta\theta_1$.

After fine-tuning, $\Delta\theta_0$ and $\Delta\theta_1$ are fixed and stored. 
When generating the intermediate image $\mathcal{I_\alpha}$, we fuse the high-level semantics in $\mathcal{I}_0$ and $\mathcal{I}_1$ by applying a linear interpolation in the low-rank parameter space:
\begin{equation}
\Delta\theta_\alpha=(1-\alpha)\Delta\theta_0+\alpha\Delta\theta_1
\end{equation}
and use the UNet with interpolated LoRA $\mathbf{\epsilon}_{\theta+\Delta\theta_\alpha}$ as the noise prediction network in the denoising steps.
Such an interpolated $\Delta\theta_\alpha$ is meaningful because $\Delta\theta_0$ and $\Delta\theta_1$ are moderately fine-tuned from the same initialization and thus are highly correlated.
While this idea of deep network interpolation~\cite{DeepNetworkInterpolation} is not new in the literature, this is the first time it has been used for image morphing with diffusion models.

\subsection{Latent Interpolation}\label{Sub:Noise}
With the noise prediction network, the next step in generating $\mathcal{I}_\alpha$ is to find the corresponding latent noise $\mathbf{z}_{T\alpha}$ and the latent text condition $\mathbf{c}_\alpha$.
To this end, we further introduce latent interpolation.

As tentatively discussed in the DDIM paper~\cite{DDIM}, a fascinating property of DDIM compared to the original DDPM~\cite{DDPM} is its suitability for image inversion and interpolation. 
Following the idea, we first get the corresponding latent noise $\mathbf{z}_{T0}, \mathbf{z}_{T1}$ for $\mathbf{z}_{00}, \mathbf{z}_{01}$ through DDIM inversion, and obtain the intermediate latent noise $\mathbf{z}_{T\alpha}$ through spherical linear interpolation (slerp)~\cite{Slerp}:
\begin{equation}
\mathbf{z}_{T\alpha} = \frac{\sin((1-\alpha)\phi)}{\sin\phi}\mathbf{z}_{T0} + \frac{\sin(\alpha\phi)}{\sin\phi}\mathbf{z}_{T1}
\end{equation}
where $\phi=\arccos \left(\cfrac{\mathbf{z}_{T0}^T \mathbf{z}_{T1}}{\Vert\mathbf{z}_{T0}\Vert\Vert\mathbf{z}_{T1}\Vert}\right)$.


However, the vanilla DDIM inversion is known to suffer from unfaithful reconstruction, especially in real image scenarios~\cite{Mokady2022NulltextIF}. 
To alleviate this, we utilize LoRA-integrated UNet $\mathbf{\epsilon}_{\theta+\Delta\theta_i} (i=0,1)$ when inverse the inputs. Since LoRA has been fine-tuned in the input images, the reconstruction from $\mathbf{z}_{Ti}$ to $\mathbf{z}_{0i}$ is much more accurate than before.

Regarding the latent text conditions $\mathbf{c}_\alpha$, we find that linear interpolations between aligned input condition $\mathbf{c}_0$ and $\mathbf{c}_1$ can serve as meaningful intermediate conditions:
\begin{equation}
\mathbf{c}_\alpha=(1-\alpha)\mathbf{c}_0+\alpha\mathbf{c}_1
\end{equation}
For example, an interpolation between ``day" and ``night" will show a gradual transition from daylight to darkness.

After getting latent noises $\mathbf{z}_{T\alpha}$ and latent condition $\mathbf{c}_\alpha$, we then denoise $\mathbf{z}_{T\alpha}$ with LoRA-integrated UNet $\mathbf{\epsilon}_{\theta+\Delta\theta_\alpha}$ using the DDIM schedule, and obtain semantically meaningful intermediate images with natural spatial transitions. 

\subsection{Self-Attention Interpolation and Replacement}\label{Sub:Attn}
Despite the semantic rationality of the intermediate results $\{\mathcal{I}_\alpha\}$, we still observe unsmooth changes in low-level textures in the generated video $\mathcal{V}$.
We attribute this problem to the highly nonlinear properties introduced in the multi-step denoising process.
To address this, we draw inspiration from attention control techniques in previous image editing studies~\cite{hertz2022prompt2prompt, Cao2023MasaCtrlTM, Parmar2023ZeroshotImageTrans, PlugAndPlay, DragDiffusion}, and propose a novel self-attention interpolation and replacement method that introduces linearly changing attention features to the denoising process and greatly reduces abrupt changes in the generated video.

\begin{figure}[t]
        \centering
        \includegraphics[width=6.5cm]{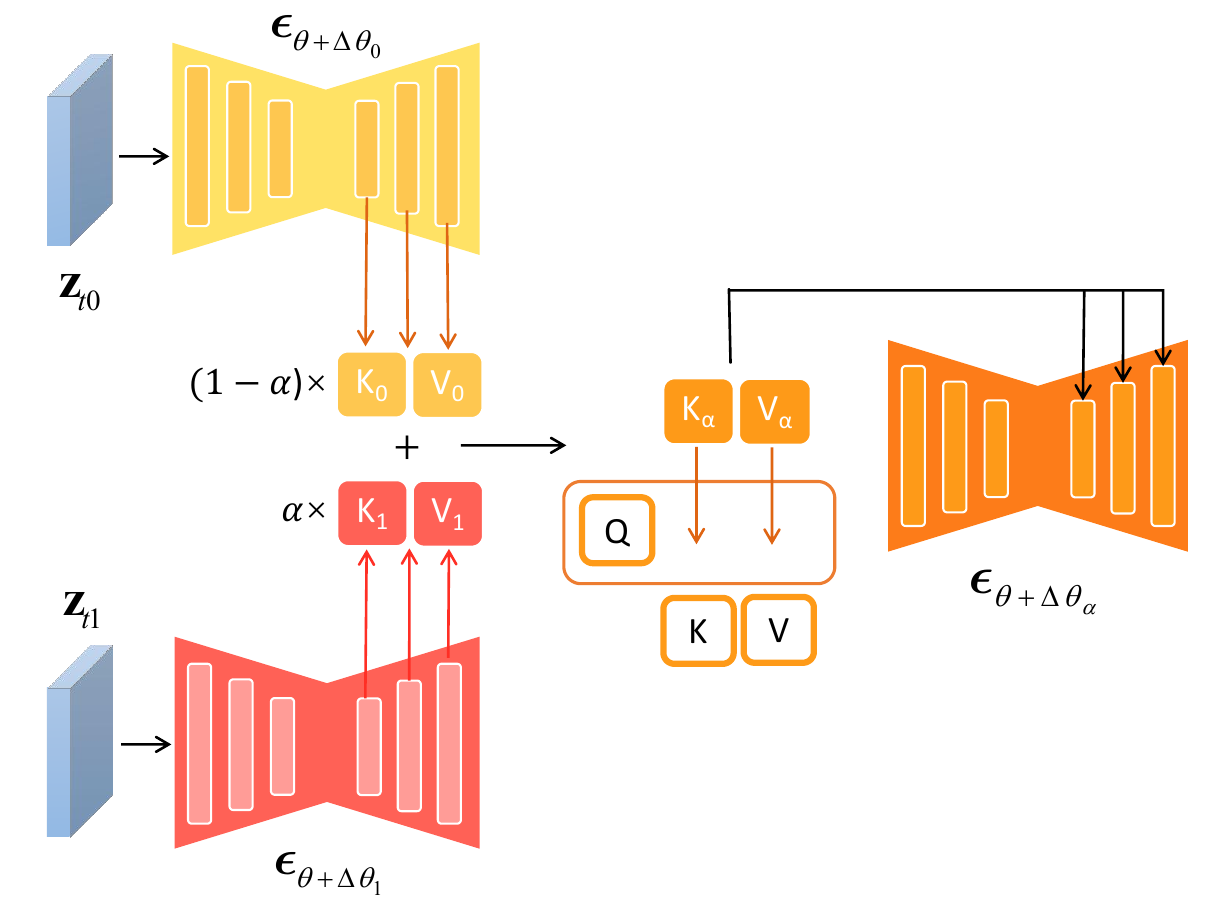}
        \vspace{-0.4cm}
	\caption{Self-Attention Interpolation and Replacement.}
        \vspace{-0.4cm}
	\label{fig:attn}
\end{figure}

As illustrated in Fig. \ref{fig:attn}, in the denoising step $t$, we first feed the latents of the input images $\mathbf{z}_{ti} (i=0,1)$ into the LoRA-integrated UNet $\epsilon_{\theta+\Delta\theta_i}$, to obtain the key and value matrices $K_i, V_i (i=0,1)$ in the self-attention modules of the UNet upsampling blocks.
In order to generate an intermediate image $\mathcal{I}_\alpha$, we linearly interpolate the matrices to get intermediate matrices:
\begin{equation}
\begin{aligned}
K_\alpha&=(1-\alpha)K_0+\alpha K_1 \\
V_\alpha&=(1-\alpha)V_0+\alpha V_1
\end{aligned}
\end{equation}
and replace the corresponding matrices in intermediate UNet $\mathbf{\epsilon}_{\theta+\Delta\theta_\alpha}$ with them.
Thus, in denoising steps, intermediate latents can query correlated local structures and textures from both input images to further enhance consistency and smoothness.

In particular, we find that replacing attention features in all denoising steps may lead to blurred image textures. Therefore, we only replace the features in the early $\lambda T (\lambda\in(0,1))$ steps and leave the self-attention modules unchanged in the remaining steps, to add high-quality details to the images. 
Empirically, we find that setting $\lambda$ to $0.4\sim 0.6$ works well in most cases.

\begin{figure*}[t]
	\centering
	\includegraphics[width=\linewidth]{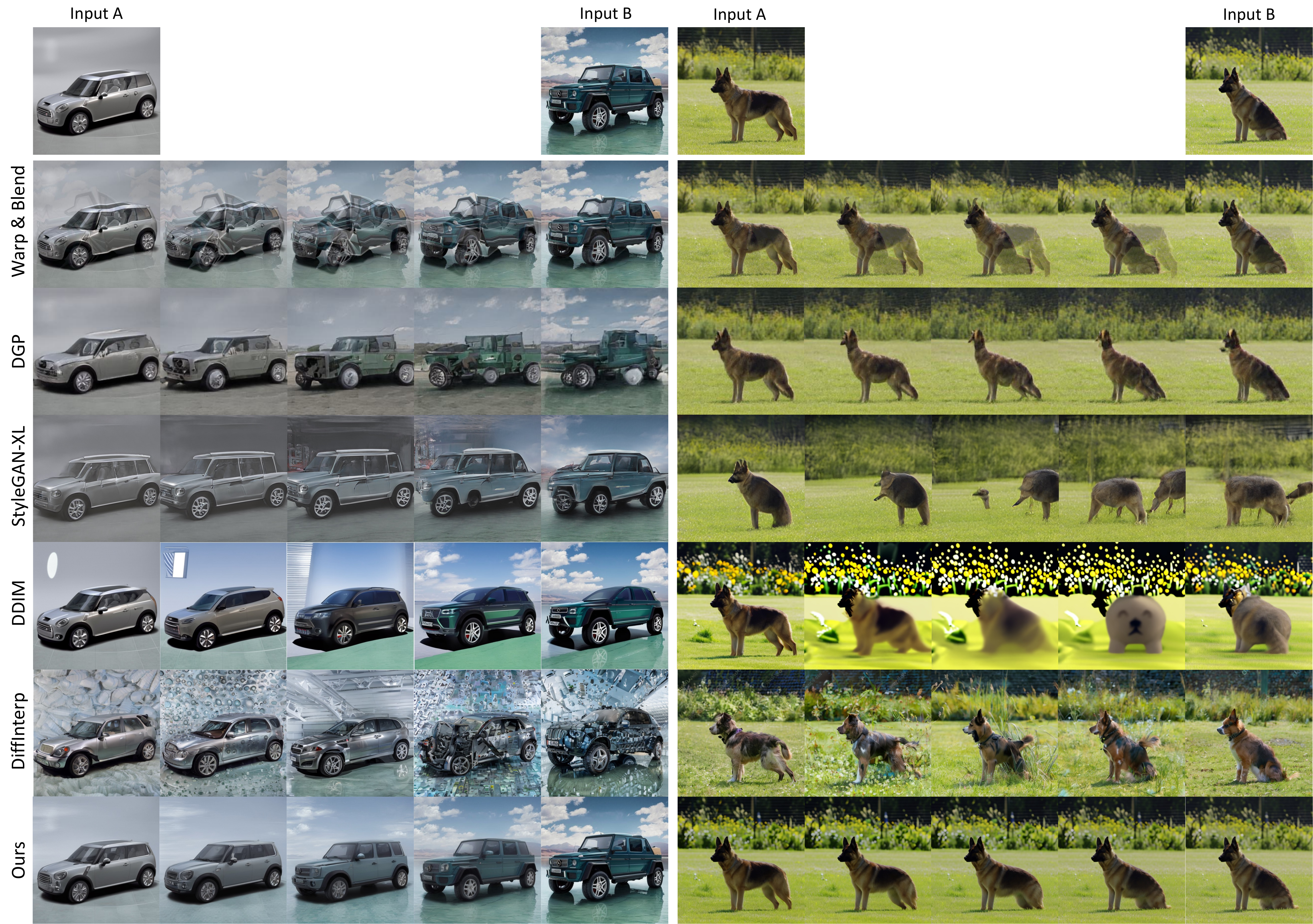}
        \vspace{-0.7cm}
	\caption{Qualitative evaluation. Our method generates intermediate images that are significantly more natural and smoother compared to those produced by previous methods. }
        \vspace{-0.3cm}
	\label{fig:qualitative}
\end{figure*}

\subsection{AdaIN Adjustment}\label{Sub:Adain}
To ensure the coherence in color and brightness between generated images and input images, we additionally introduce the Adaptive Instance Normalization (AdaIN)~\cite{adain} adjustment for interpolated latent noise $\mathbf{z}_{0\alpha} (\alpha \in (0,1))$ before denoising.

Specifically, we calculate the mean $\mu_i$ and standard deviation $\sigma_i (i=0,1)$ for each channel of latent noises $\mathbf{z}_{00}, \mathbf{z}_{01}$, and interpolate between $\mu_i, \sigma_i$ as the adjustment target of intermediate noises:
\begin{align}
    \mu_\alpha &= (1-\alpha)\mu_0 + \alpha\mu_1 \\
    \sigma_\alpha &= (1-\alpha)\sigma_0 + \alpha\sigma_1 \\
    \tilde{\mathbf{z}}_{0\alpha} &= \sigma_\alpha \left( \cfrac{\mathbf{z}_{0\alpha}-\mu(\mathbf{z}_{0\alpha})}{\sigma(\mathbf{z}_{0\alpha})} \right) + \mu_\alpha
\end{align}
and replace the intermediate latent noise $\mathbf{z}_{0\alpha}$ with the adjusted one $\tilde{\mathbf{z}}_{0\alpha}$ in the denoising process.
As demonstrated in Fig.~\ref{fig:adain}, the color and brightness are more coherent after AdaIN adjustment. 

\begin{figure*}[t]
	\centering
	\includegraphics[width=\linewidth]{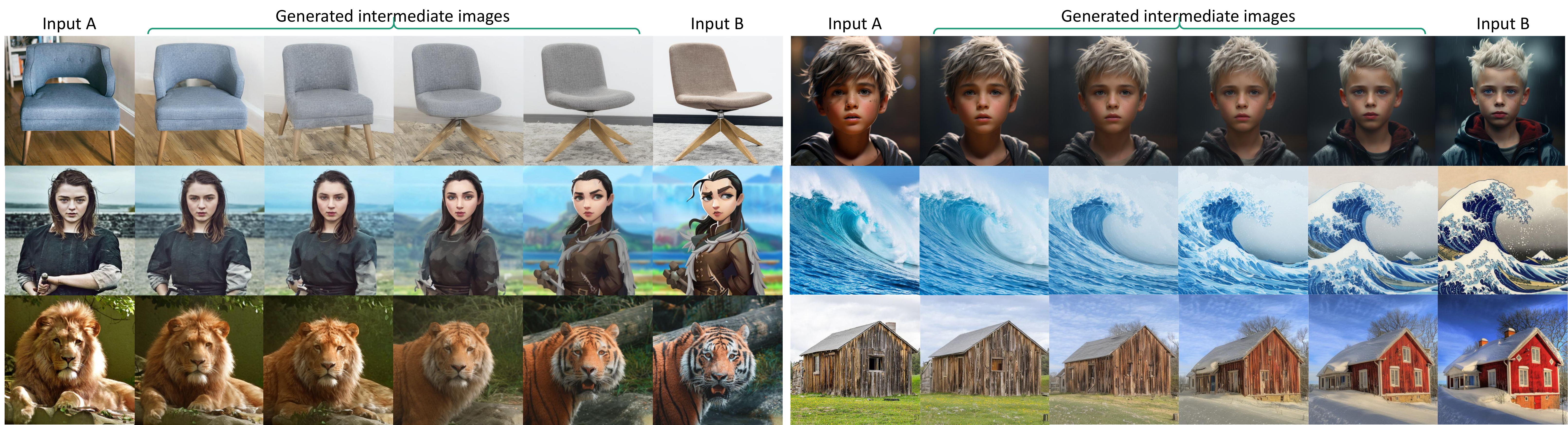}
        \vspace{-0.7cm}
	\caption{Qualitative results. Our approach achieves visually pleasing image morphing across diverse object categories. }
        \vspace{-0.2cm}
	\label{fig:qualitative2}
\end{figure*}

\subsection{Reschedule Sampling}\label{Sub:Reschedule}
With all the methods introduced above, we can generate a smooth transition video between two input images with natural and high-quality in-betweens.
However, we observe that using a naive linear sampling schedule for $\alpha$ may result in an uneven transition rate in image content.
To achieve a homogeneous transition rate, we further introduce a new reschedule method.

\begin{figure*}[t]
	\centering
	\includegraphics[width=\linewidth]{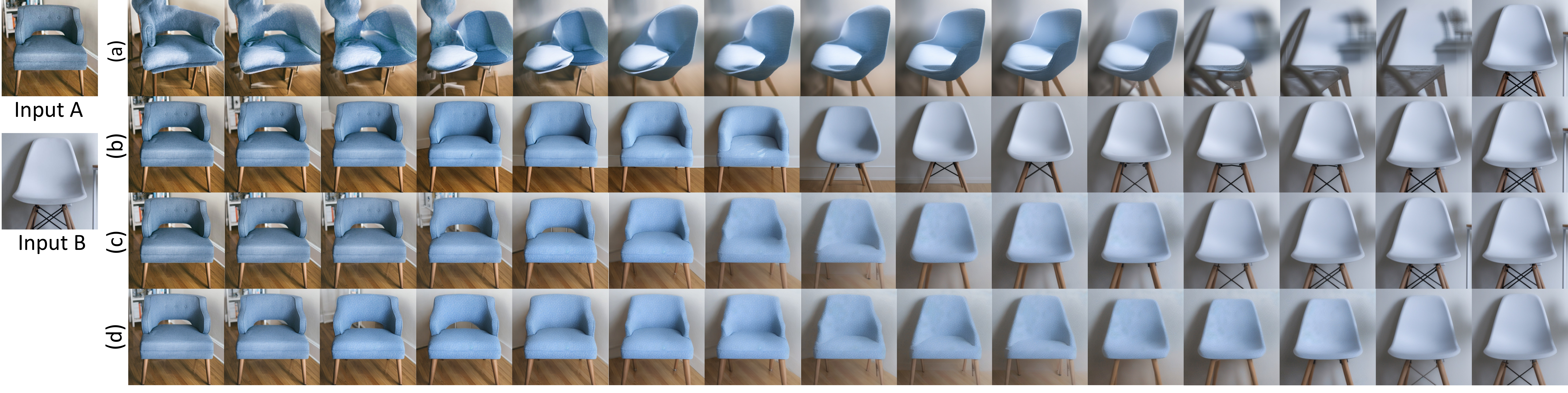}
        \vspace{-0.75cm}
	\caption{Ablation study. The four settings are the same as in Table~\ref{tab:ablation}: (a) DDIM baseline, (b) + LoRA interpolation, (c) + attention interpolation and replacement, (d) + reschedule (Ours). }
	\label{fig:ablation}
\end{figure*}

Formally, assuming $D(\mathcal{I}_{i}, \mathcal{I}_{j}) (i,j \in [0, 1])$ is the perceptual distance between $\mathcal{I}_{i}$ and $\mathcal{I}_{j}$, given the number of frames $n$, we want the variance of $\{D(\mathcal{I}_{i}, \mathcal{I}_{i + \frac{1}{n}}) | i = 0, \frac{1}{n}, \cdots, 1 - \frac{1}{n} \}$ to be as small as possible. The reschedule sampling starts with approximating the gradient of the relative perceptual distance $\Delta D$ with respect to $\alpha$:
\begin{equation}
\Delta D (\alpha) =
\begin{cases} 
D(\mathcal{I}_{0}, \mathcal{I}_{\frac{1}{n}}) / \bar{D} & \text{if } 0 \leq \alpha < \frac{1}{n}, \\
D(\mathcal{I}_{\frac{1}{n}}, \mathcal{I}_{\frac{2}{n}}) / \bar{D} & \text{if } \frac{1}{n} \leq \alpha < \frac{2}{n}, \\
\vdots & \vdots \\
D(\mathcal{I}_{1 - \frac{1}{n}}, \mathcal{I}_{1}) / \bar{D} & \text{if } 1 - \frac{1}{n} \leq \alpha \leq 1,
\end{cases}
\end{equation}
where $\bar{D} = \sum_{i=0}^{1 - \frac{1}{n}} D(\mathcal{I}_{i}, \mathcal{I}_{i + \frac{1}{n}})$ is the sum of perceptual distance between all adjacent frames. Then the relative perceptual distance to the first frame for every  $\alpha$ can be estimated with 
\begin{equation}
D_0 (\alpha) = \int_{0}^{\alpha} \Delta D(x) dx.
\end{equation}
Finally, by utilizing $D_0$ and its inversion function $D_0'$, we can determine the rescheduled interpolation parameters $\alpha_i$ as $\{\alpha_i = D_0'(y) | y = 0, \frac{1}{n}, \cdots, 1\}$. As demonstrated in Fig. \ref{fig:ablation}, the new sampling schedule ensures a more uniform transition rate in the image content.

\section{Experiment}
\label{Section:Exp}

\begin{table*}[t]
	\centering
	\begin{minipage}{0.55\linewidth}
	\caption{Quantitative evaluation on \textit{MorphBench}. We report FID ($\downarrow$), PPL ($\downarrow$), and perceptual distance variance (PDV, $\downarrow$) to evaluate the fidelity, smoothness, and speed homogeneity of the transition video respectively.}
        \vspace{-0.3cm}
	\resizebox{\linewidth}{!}{
		\begin{tabular}{l|ccc|ccc|ccc}
			\hline
			\multirow{2}{*}{Method} & \multicolumn{3}{c|}{Metamorphosis} & \multicolumn{3}{c|}{Animation} & \multicolumn{3}{c}{Overall} \\
			& FID    & PPL    & PDV    & FID      & PPL      & PDV     & FID    & PPL    & PDV    \\ \hline\hline
			Warp \& Blend &     79.63   &   15.97     &  4.64      &     56.86     &     9.58     &    0.99     &   67.57     &    14.27    &    3.67    \\
			DGP                &    150.29    &    29.65    &   79.40     &     194.65     &    27.50      &     34.21    &     138.20   &   29.08     &  67.35      \\
			StyleGAN-XL\hspace{-4pt} &    122.42    &  41.94      &    181.50    &    133.73      &   33.43       &     37.95    &   112.63     &    39.67    &  143.22      \\
			DDIM              &     95.44   &   27.80     &    302.83    &    174.31      &    18.70      &  249.16       &      101.68  &    25.38    &   288.51     \\
			Diff.Interp.       &     169.07   &    108.51    &    135.95    &     148.95     &     96.12     &    49.27     &     146.66   &    105.23    &  112.84      \\
			Ours               &    70.49    &    18.19    &  22.93      &     43.15     &     5.14     &    3.55     &     54.69   &    21.10    &    21.42 \\ \hline
		\end{tabular}
		\label{tab:quant}
	}
	\end{minipage}
        ~
	\begin{minipage}{0.435\linewidth}
		\centering
		\caption{Ablation study. We study the effects of each proposed component in our method.}
		\resizebox{\linewidth}{!}{
			\begin{tabular}{c|ccc|ccc}
				\hline
				Method & \makecell{LoRA \\ Interp. }\hspace{-6pt}  &  \makecell{Attention \\ Interp.}\hspace{-6pt}  &  \makecell{AdaIN \\ \& Reschedule} \hspace{-6pt}  & FID & PPL & PDV   \\ \hline\hline
				DDIM        &       &         &         &  101.68  &  25.38    &   288.51  \\
				-   &  \checkmark    &         &        &  
                44.40   &  21.81 & 249.33    \\
				-   &  \checkmark    &    \checkmark  &     &
                44.90  &  19.86  &  157.73      \\
				Ours   &  \checkmark    &    \checkmark  & \checkmark & 54.69  & 21.10  & 21.42   \\  \hline
			\end{tabular}
			\label{tab:ablation}
		}
	\end{minipage}
        \vspace{-0.3cm}
\end{table*}

\subsection{Implementation Details}
In all of our experiments, we use the publicly available state-of-the-art Stable Diffusion v2.1-base as our diffusion model.
When training LoRA, to achieve a balance between efficiency and quality and avoid overfitting the single image, we only fine-tune the projection matrices $Q, K, V$ in the attention modules of the diffusion UNet.
Additionally, we set the rank of LoRA to 16, and train for 200 steps using AdamW optimizer~\cite{AdamW} with a learning rate of $2\times 10^{-4}$.
In this setting, training a LoRA for a $512\times 512$ image requires only $\sim 20$s on a NVIDIA A100 GPU.

During the inversion and denoising process, we adopt the DDIM schedule of 50 steps distilled from entire diffusion steps $T=1000$.
It's noteworthy that we do not apply classifier-free guidance (CFG)~\cite{CFG} in both DDIM inversion and denoising. This is because CFG tends to accumulate numerical errors and cause supersaturation problems, which is also observed in \cite{Mokady2022NulltextIF, DragDiffusion}.
For attention control, we only perform the feature injection in the upsampling blocks in the self-attention module of the diffusion UNet, and set the hyperparameter $\lambda$ to 0.6 by default.

\subsection{\textbf{MorphBench}}
Conventional image morphing techniques in computer graphics generally require tedious manual labeling of correspondences, and general image morphing is rarely explored in depth in the area of generative models.
Therefore, there is a lack of specific evaluation benchmarks for this task.
To comprehensively evaluate the effectiveness of our methods, we present \textit{MorphBench}, the first benchmark dataset for assessing image morphing of general objects.
We collect 90 pairs of pictures of diverse content and styles, and divide them into two categories: i) \textit{metamorphosis} between different objects (66 pairs) and ii) \textit{animation} of the same objects (24 pairs).
The latter is obtained using off-the-shelf image editing tools such as DragDiffusion~\cite{DragDiffusion}, Imagic~\cite{Imagic}, and MasaCtrl~\cite{Cao2023MasaCtrlTM}.
We hope \textit{MorphBench} can also promote future studies on this important problem.

\subsection{Qualitative Evaluation}
To demonstrate the superiority of our methods, we provide a visual comparison of the results produced by the previous methods and ours.
We extensively compare our outcomes with five representative image morphing methods for general objects, including the following three types: i) classic graphical morphing technique~\cite{Triangle} based on warping and blending; 
ii) GAN-based deep interpolation methods DGP~\cite{DGP} and StyleGAN-XL~\cite{StyleGANXL} trained on large-scale general image dataset~\cite{Deng2009ImageNetAL}; 
iii) diffusion-based deep interpolation methods DDIM~\cite{DDIM} and Diff.Interp.~\cite{wang2023interpolating} based on Stable Diffusion v2.1-base~\cite{LDM}. 
More details about the baselines we use can be found in the supplementary material.

As demonstrated in Fig. \ref{fig:qualitative}, our \textit{DiffMorpher} outperforms all previous approaches in terms of image fidelity, semantic consistency, and transformation smoothness, whether used to morph between different objects or animate the same object.
We observe that previous approaches suffer from artifacts of different characteristics, while the results of our method are much more visually pleasing.
More qualitative results are presented in Fig. \ref{fig:qualitative2}.
With a single diffusion model, our approach well handles diverse object categories and image styles.
It is worth mentioning that the generated images accurately reflect the dense correspondence between the two input images, although no such annotation is provided.
We recommend readers refer to the supplementary material for video results. 

\subsection{Quantitative Evaluation}
To quantitatively evaluate the quality of intermediate images and the smoothness of the transition video, we follow the metrics adopted in the baseline Diff.Interp.~\cite{wang2023interpolating}:

(1) Frechet inception distance (FID, $\downarrow$)~\cite{FID}: We compute the FID score between the distribution of the input images and the distribution of the generated images.
To estimate the distribution of generated images, we randomly sample two images from the interpolation video 10 times and calculate the mean FID score as an index of the rationality and fidelity of intermediate images.

(2) Perceptual path length (PPL, $\downarrow$)~\cite{PPL}: We compute the sum of the perceptual loss~\cite{zhang2018perceptual} between adjacent images in 17-frame sequences, as an index of the smoothness and consistency of the transition video.

Furthermore, in order to measure the homogeneity of the video transition rate, we introduce a new metric:

(3) Perceptual distance variance (PDV, 
$\downarrow$): We compute the perceptual loss between consecutive images in 17-frame sequences just like PPL, and then calculate the variance of these distances in the sequence. 
The average distance variance of all sequences from the test set is taken as the PDV index.
This can be a natural measurement of the homogeneity of the video transition rate, where a lower PDV indicates a more uniform speed.

The quantitative results of all approaches are presented in Table~\ref{tab:quant}.
Our approach achieves significantly lower FID in both \textit{metamorphosis} and \textit{animation} scenarios, showing better image fidelity and consistency with the input images.
Although the classic Warp \& Blend approach shows better PPL and PDV scores, this is due to the smooth and linear nature of the warping and blending operation which is prone to ghosting artifacts as can be seen in Fig.~\ref{fig:qualitative}.
Among all the deep interpolation methods, our approach has far lower PPL and PDV than others, demonstrating smoother transition video and more homogeneous speed of content change.
These results are consistent with the qualitative comparison.

\subsection{Ablation Study}

\begin{figure}[t]
        \includegraphics[width=\linewidth]{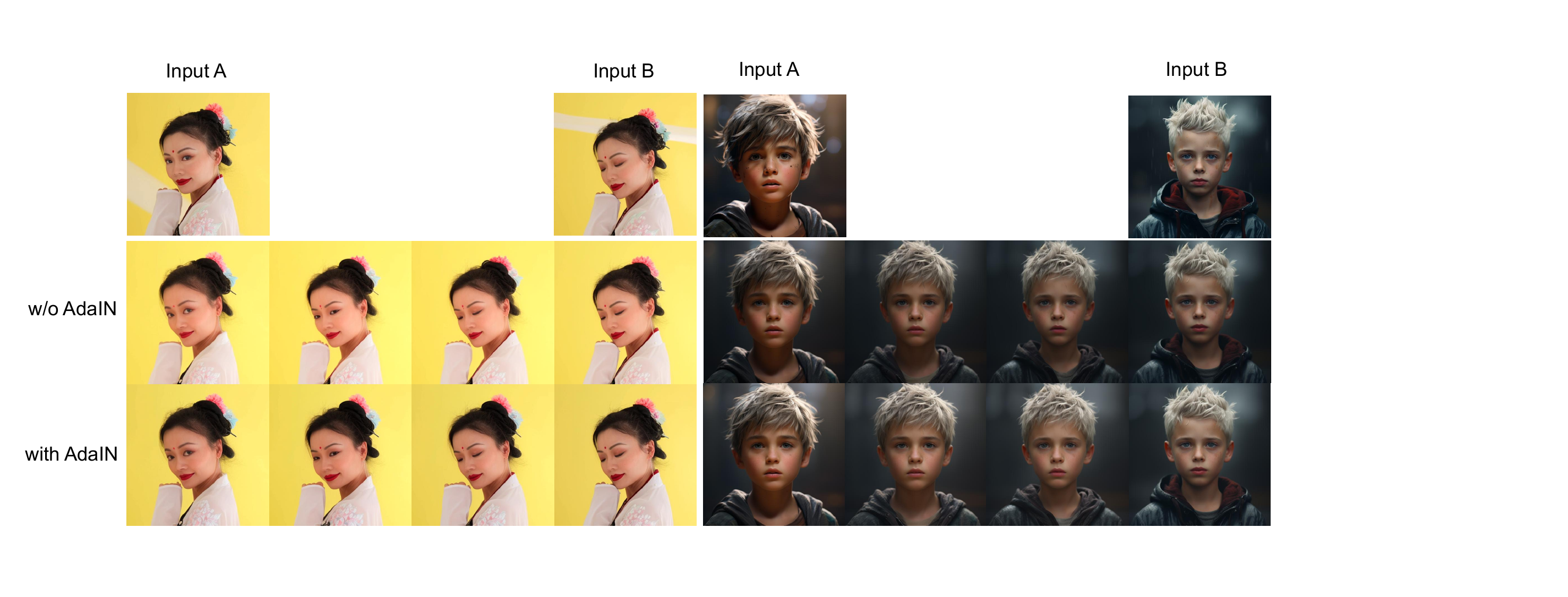}
        \vspace{-0.8cm}
	\caption{Effects of AdaIN adjustment. The colors and brightness of the intermediate images become more consistent with the input images after AdaIN adjustment.}
        \vspace{-0.2cm}
	\label{fig:adain}
\end{figure}

To verify the effectiveness of each proposed component, we perform an ablation study and show the results in Table~\ref{tab:ablation} and Fig.~\ref{fig:ablation}.
The most critical component is LoRA interpolation, which fixes the corrupted images of DDIM to be high-fidelity and semantically smooth images, thus reducing FID, PPL, and PDV.
However, abrupt content changes can still be observed, such as the 7th and 8th images of Fig.~\ref{fig:ablation} (b).
The attention interpolation and replacement technique effectively eliminates such abrupt changes and makes the image sequence much smoother as shown in Fig.~\ref{fig:ablation} (c), further improving PPL and PDV.
Despite so, the speed of content change is still uneven, \eg, the first three images or the last three images of Fig.~\ref{fig:ablation} (c) are almost the same while the content change during 7-9th images is much faster.
As shown in Fig.~\ref{fig:ablation} (d), this problem is addressed with our new sampling schedule, which redistributes the content change to be balanced among all consecutive images and thus cuts down PDV by a large margin.
Note that this leads to slightly higher FID, because results without rescheduling are biased toward the two ends and thus are closer to the two input images.
Lastly, after applying AdaIN adjustment to the latent noises, 
the colors and brightness are more consistent than before, as shown in Fig.~\ref{fig:adain}.

\begin{figure}[t]
	\centering
	\includegraphics[width=\linewidth]{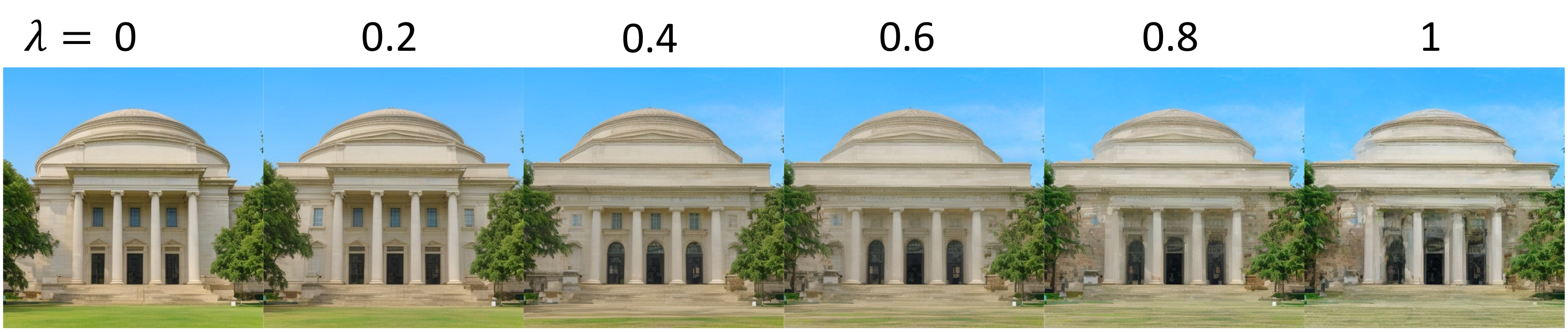}
        \vspace{-0.7cm}
	\caption{Effects of $\lambda$. We show an intermediate image of the second example in Fig.~\ref{fig:teaser} with different $\lambda$. The image starts to get blurry when $\lambda > 0.6$. }
	\label{fig:lambda}
        \vspace{-0.2cm}
\end{figure}

In our method, $\lambda$ is used to control the strength of attention replacement.
We further study its effects in Table~\ref{tab:lambda} and Fig.~\ref{fig:lambda}.
As $\lambda$ increases, more attention replacement is involved in the denoising steps, thus improving smoothness and reducing PPL and PDV.
However, using interpolated attentions in the later denoising steps can harm the generation of low-level textures and blurry artifacts may emerge, as demonstrated in Fig.~\ref{fig:lambda}.
We found that setting $\lambda=0.6$ achieves a good balance between video smoothness and image quality.

\begin{table}[t]
	\centering
	\caption{Effects of $\lambda$.}
        \vspace{-0.3cm}
	\resizebox{0.75\linewidth}{!}{
		\begin{tabular}{ccccccc}
			\hline
			$\lambda$ & 0  &  0.2  &  0.4 & 0.6 & 0.8 & 1   \\ \hline\hline
		    FID     &    53.78     & 52.99  & 53.45      & 54.69   & 52.47  &  55.84   \\
            PPL     &    23.85     &  23.25      &  22.26    &  21.10  &  19.49 & 17.85   \\
			PDV    &    81.15     &   62.64      &   36.26     &  21.42    & 15.79  &  12.86  \\  \hline
		\end{tabular}
		\label{tab:lambda}
	}
        \vspace{-0.4cm}
\end{table}


\section{Conclusion}

We have presented \textit{DiffMorpher}, an image morphing approach that only relies on the prior knowledge of a pre-trained text-to-image diffusion model.
Our method is able to generate a sequence of visually pleasing images that deliver a smooth transition between two input images.
This is achieved by capturing the semantics of the two images via two LoRAs, and interpolating in both the LoRA parameter space and the latent noise to produce a smooth semantic interpolation.
An attention interpolation and injection method, an AdaIN adjustment technique, and a new sampling schedule are further introduced to motivate smoothness between consecutive images.
We have demonstrated that our approach significantly advances the state of the art in image morphing, 
uncovering the large potential of diffusion models in this task.

{\small
\bibliographystyle{ieee_fullname}
\bibliography{egbib}
}
\newpage
\setcounter{section}{0}
\renewcommand\thesection{\Alph{section}}
\section{More Details of Baselines}
In Sec.~\ref{Section:Exp}, we comprehensively compare our method with previous state-of-the-art methods, including graphical, GAN-based and diffusion-based techniques.
We offer more details of the baselines that we use here:
\begin{itemize}
    \item Warp \& Blend~\cite{Survey98, Survey17, Survey23}: Conventional graphical techniques usually involve bidirectional image warping based on correspondence point pairs with blending operations to achieve morphing effects.
    We select the representative triangulation-based method~\cite{Triangle} as our baseline, which is also widely used in standard libraries such as OpenCV.
    It divides the images into triangles by performing Delaunay triangulation on user-defined corresponding points, and then morphs between the triangle pairs.
    Thus, the quantity and quality of the manually labeled pair of points greatly affect the generated results.
    Since all the other methods do not require correspondence annotations, for the sake of fairness, we adopt the automatic version of this approach \href{https://github.com/jankovicsandras/autoimagemorph}{https://github.com/jankovicsandras/autoimagemorph} that selects 50 morph-points automatically using OpenCV.
    
    \item Deep Generative Prior (DGP)~\cite{DGP}: DGP is an image manipulation method based on BigGAN~\cite{BigGAN}, which is suitable for general image morphing.
    We adopt the official code \href{https://github.com/XingangPan/deep-generative-prior}{https://github.com/XingangPan/deep-generative-prior} with its default hyperparameters and the pretrained BigGAN model trained on ImageNet~\cite{BigGAN} as our baseline.

    \item StyleGAN-XL~\cite{StyleGANXL}: Since the pretrained checkpoint of StyleGAN-T~\cite{StyleGANT} is not publicly available, we use the alternative state-of-the-art GAN model StyleGAN-XL \href{https://github.com/autonomousvision/stylegan-xl}{https://github.com/autonomousvision/stylegan-xl} as our another baseline.
    Similarly to DGP, the model is trained on ImageNet.
    We obtain the latent codes of input images by GAN inversion~\cite{GANInversion} and tune the generator by PTI~\cite{PTI} for better reconstruction results, and interpolate both the latent codes and the generator parameters to get intermediate images.
    For both GAN-based methods, we use the ImageNet classifier DeiT~\cite{deit} to automatically determine the class label.

    \item Denoising Diffusion Implicit Model (DDIM)~\cite{DDIM}: We implement a naive diffusion-based interpolation method through DDIM inversion and latent interpolation as our baseline, as discussed in the DDIM paper.
    As with our approach, the underlying model used is also Stable Diffusion v2.1-base \href{https://huggingface.co/stabilityai/stable-diffusion-2-1-base}{https://huggingface.co/stabilityai/stable-diffusion-2-1-base}.

    \item Diff.Interp.~\cite{wang2023interpolating}: \textit{Interpolating between Images with Diffusion Models} is a recent state-of-the-art image interpolation method based on diffusion models.
    Besides latent interpolation, it further introduced pose guidance based on ControlNet~\cite{controlnet} to encourage more reasonable intermediate results.
    However, the smoothness of the morphing video was not considered in this work, and the generated video is full of flickering artifacts.
    We employ the official code \href{https://github.com/clintonjwang/ControlNet}{https://github.com/clintonjwang/ControlNet} with default settings  and pretrained Stable Diffusion v2.1-base model as our baseline.
    For all three diffusion-based methods, the prompts for each test case are shared.
\end{itemize}

\section{User Study}
\begin{figure}[t]
        \includegraphics[width=8.5cm]{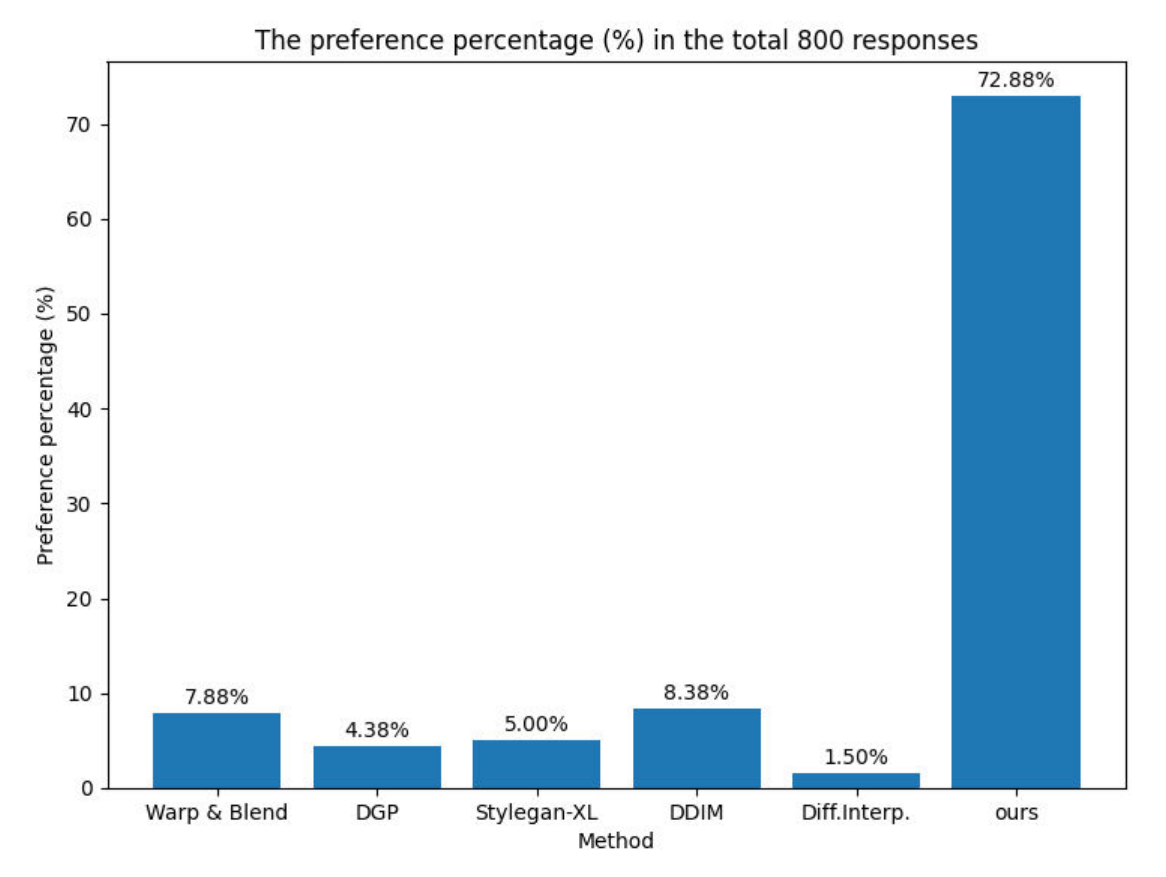}
        \vspace{-0.8cm}
	\caption{User study result. Our method surpasses all the previous methods by a large margin in terms of user preference.}
        \vspace{-0.4cm}
	\label{fig:userstudy}
\end{figure}

To assess the quality of image morphing from a human perspective, we invite 40 volunteers to conduct a user study.
Each participant are shown 20 groups of morphing videos created by our approach and five baseline methods, chosen at random.
They are asked to evaluate the image morphing quality from the perspective of intermediate image fidelity and video smoothness, and to select the one with the best quality for each question.
An example of the questionnaire is shown in Fig.~\ref{fig:question}.
In total, we collect 800 responses and summarize the results in Fig.~\ref{fig:userstudy}.
As we can see, our approach is significantly more preferred by users than any of the prior methods.

\section{Limitations}
\begin{figure}[t]
        \includegraphics[width=8.5cm]{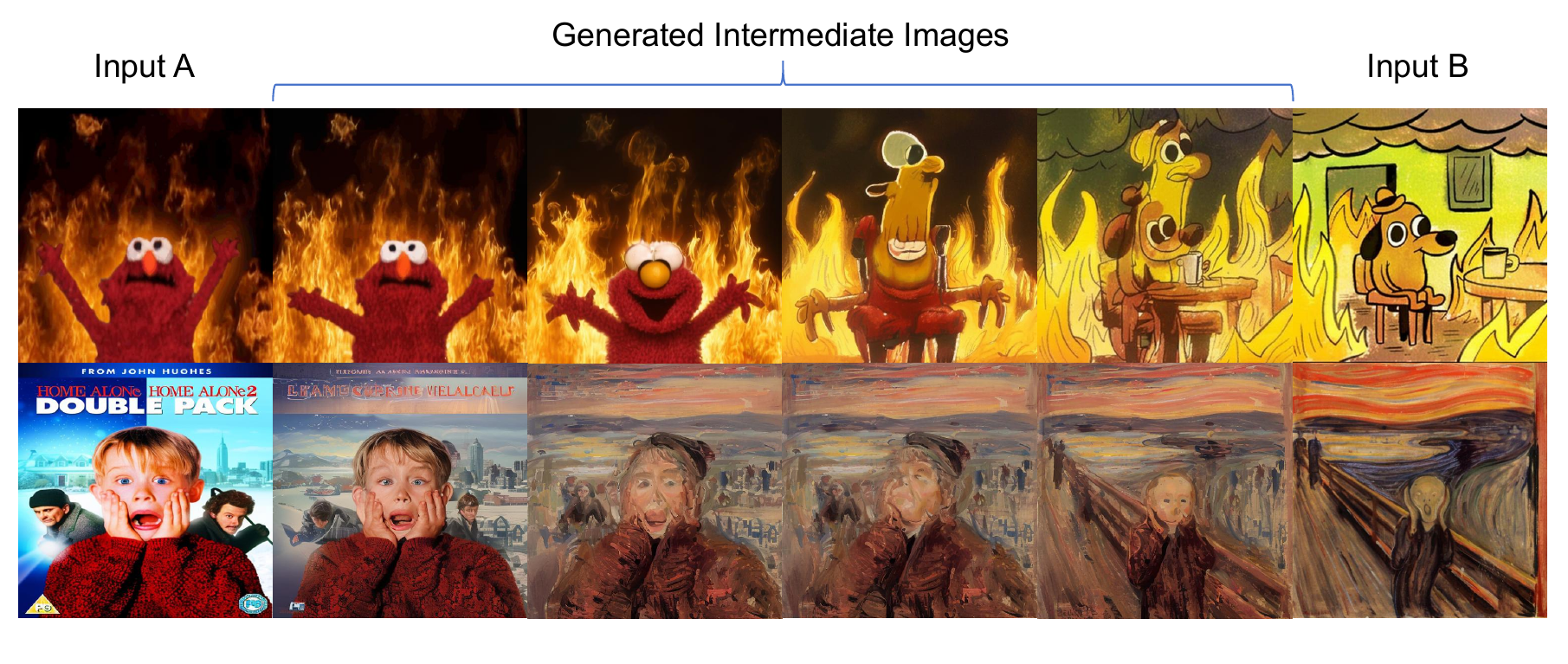}
        \vspace{-0.8cm}
	\caption{Some relatively unsuccessful cases where the correspondence between two images is not clear enough.}
        \vspace{-0.4cm}
	\label{fig:failcase}
\end{figure}

One of the limitations of our approach is that we have to train a LoRA for each input image before morphing, which costs additional time ($\sim 20$ s on a single NVIDIA A100 GPU for a $512\times 512$ image).
Another limitation of text-guided diffusion models is that the user must input aligned text prompts in addition to images.
Besides, our approach occasionally fails in difficult cases where the correspondence between two input images is not clear enough, and produces relatively unreasonable intermediate images, as shown in Fig.~\ref{fig:failcase}.

\section{More Qualitative Results}
\begin{figure*}[t]
	\centering
	\includegraphics[width=\linewidth]{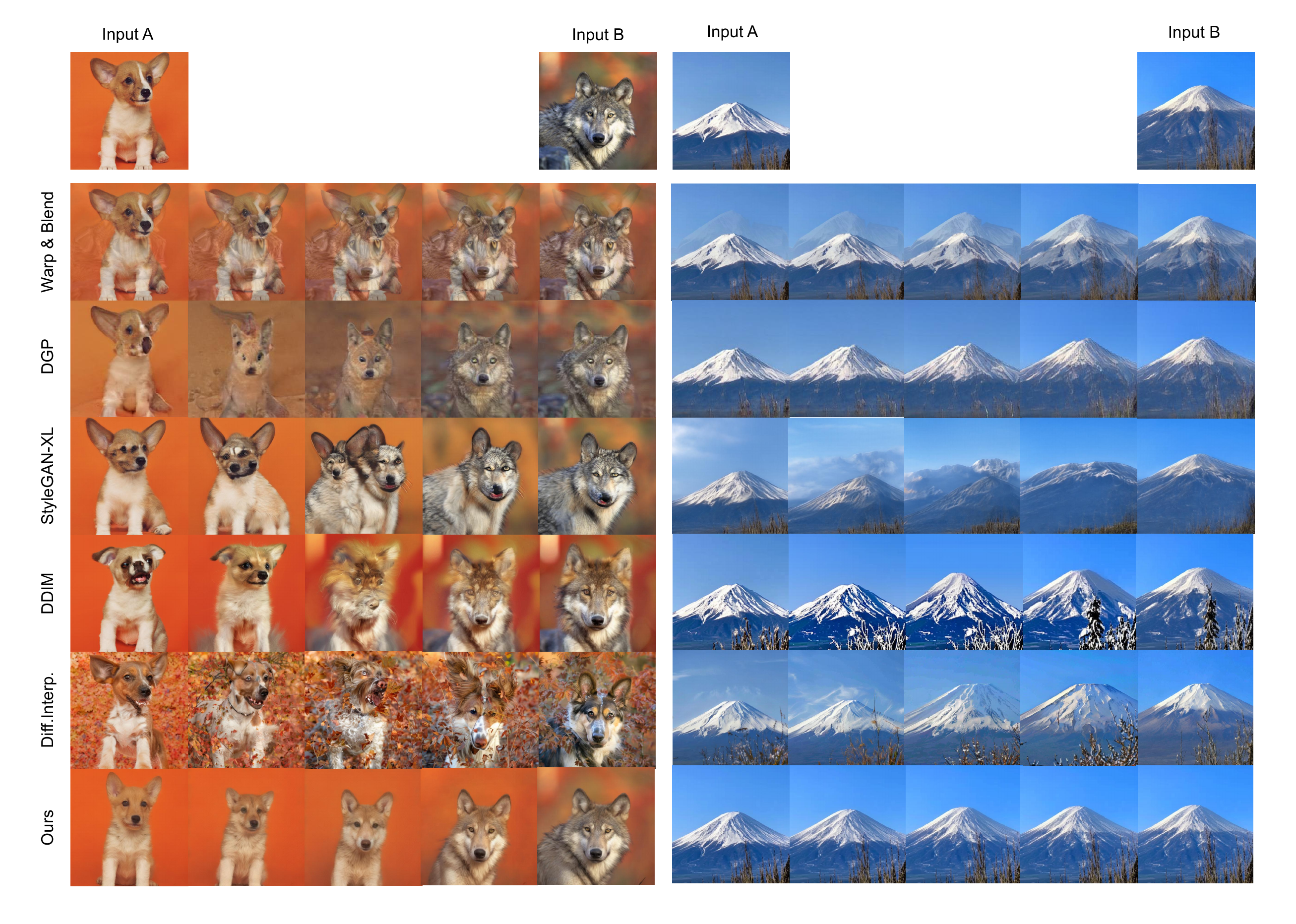}
        \vspace{-0.7cm}
	\caption{More qualitative comparison results. }
        \vspace{-0.3cm}
	\label{fig:compare}
\end{figure*}

Here we present more qualitative results to demonstrate the effectiveness of our \textit{DiffMorpher}.
Fig.~\ref{fig:compare} gives more examples to illustrate the superiority of our approach compared to previous methods in diverse scenarios, and
Fig.~\ref{fig:ours} and Fig.~\ref{fig:ours2} provide additional qualitative results generated by our method that further demonstrate its versatility in real-world applications.

\twocolumn[{
	\renewcommand\twocolumn[1][]{#1}
	\maketitle
	\vspace{-10mm}
	\begin{center}
		\includegraphics[width=0.96\textwidth]{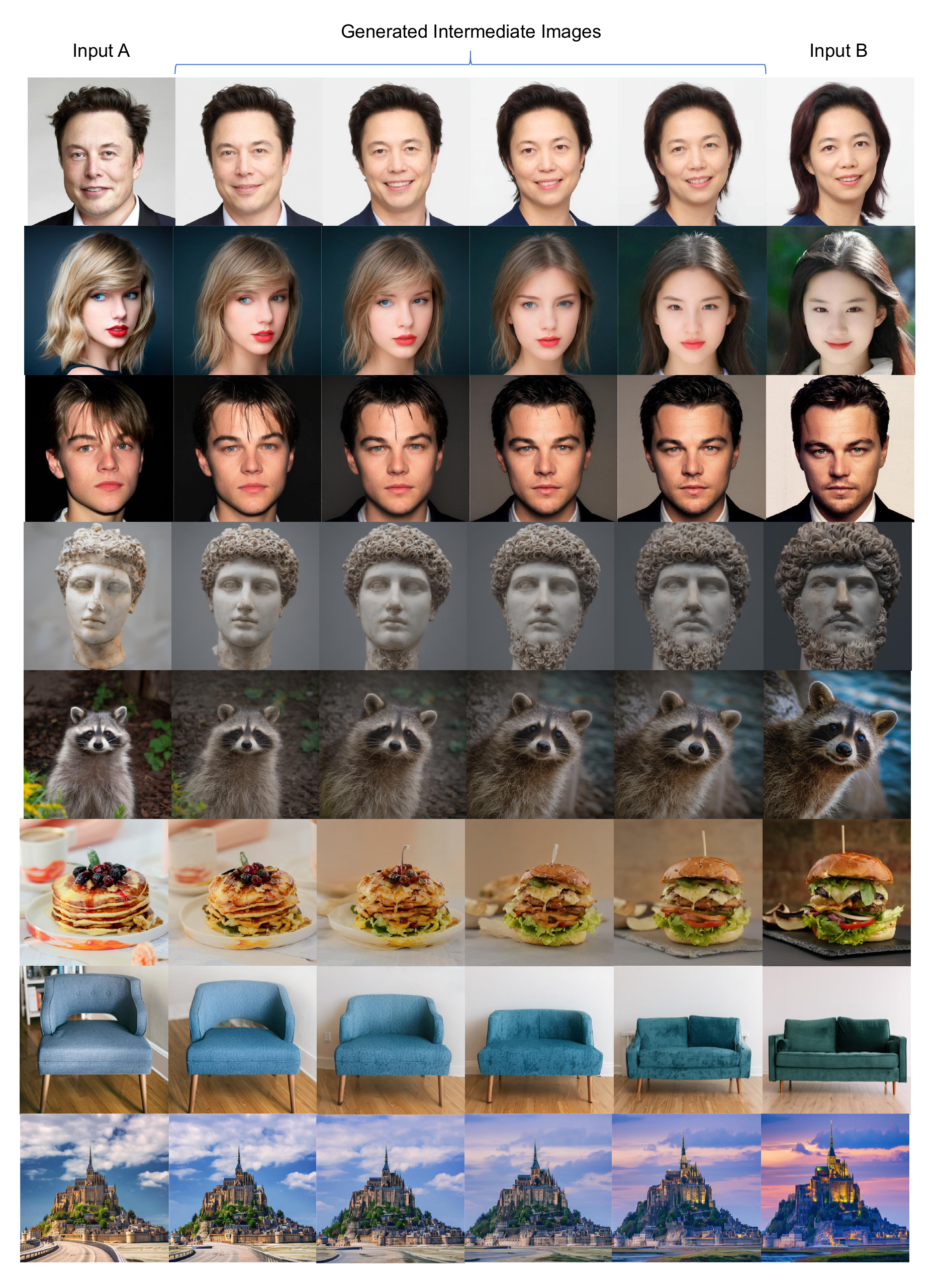}
	\end{center}
	\vspace{-6mm}
	\captionof{figure}{More qualitative results of our approach.}
	\label{fig:ours}
	\vspace{6mm}
}]

\twocolumn[{
	\renewcommand\twocolumn[1][]{#1}
	\maketitle
	\vspace{-10mm}
	\begin{center}
		\includegraphics[width=0.96\textwidth]{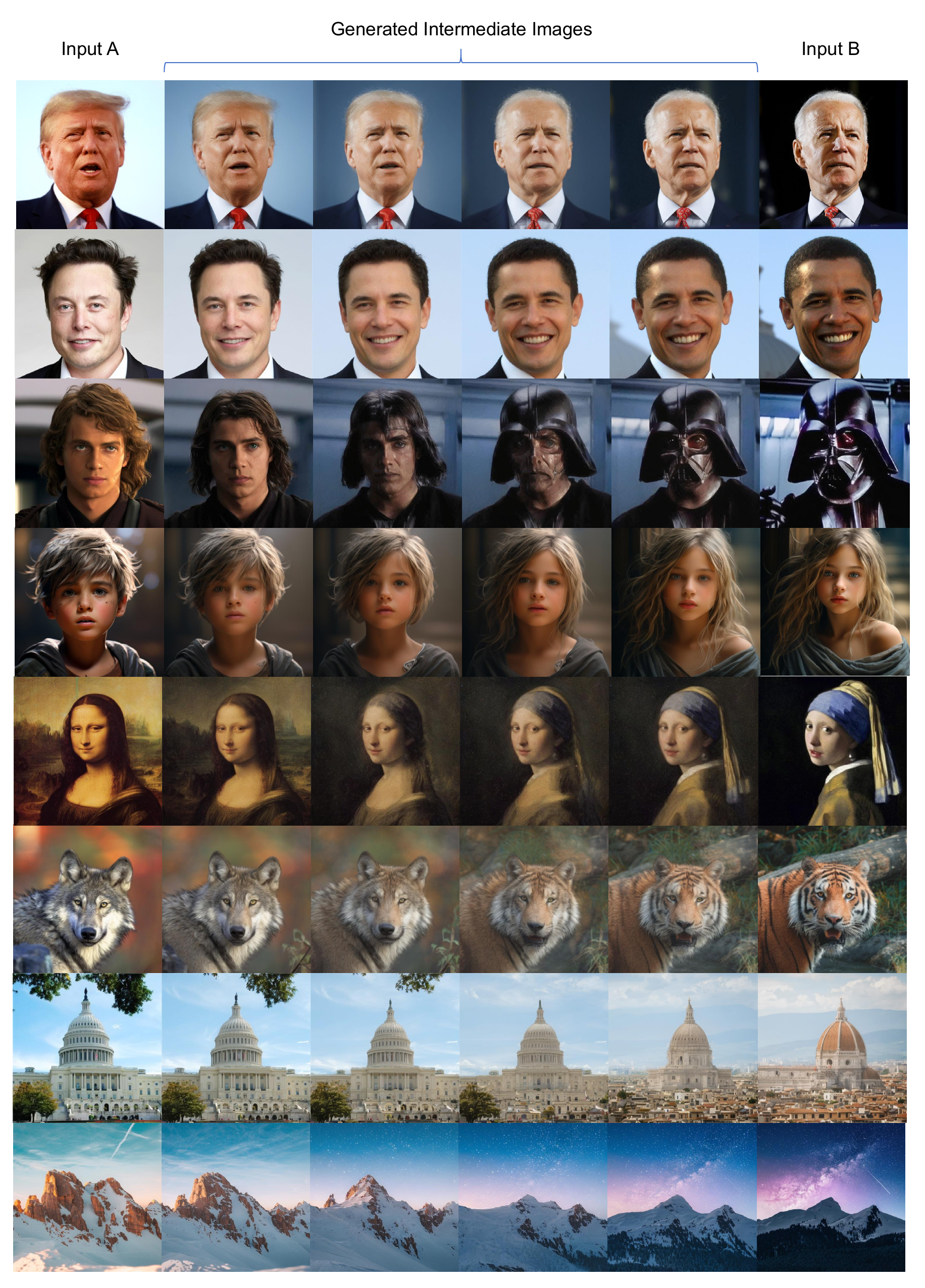}
	\end{center}
	\vspace{-6mm}
	\captionof{figure}{More qualitative results of our approach.}
	\label{fig:ours2}
	\vspace{6mm}
}]

\twocolumn[{
	\renewcommand\twocolumn[1][]{#1}
	\maketitle
	\vspace{-7mm}
	\begin{center}
		\includegraphics[width=0.96\textwidth]{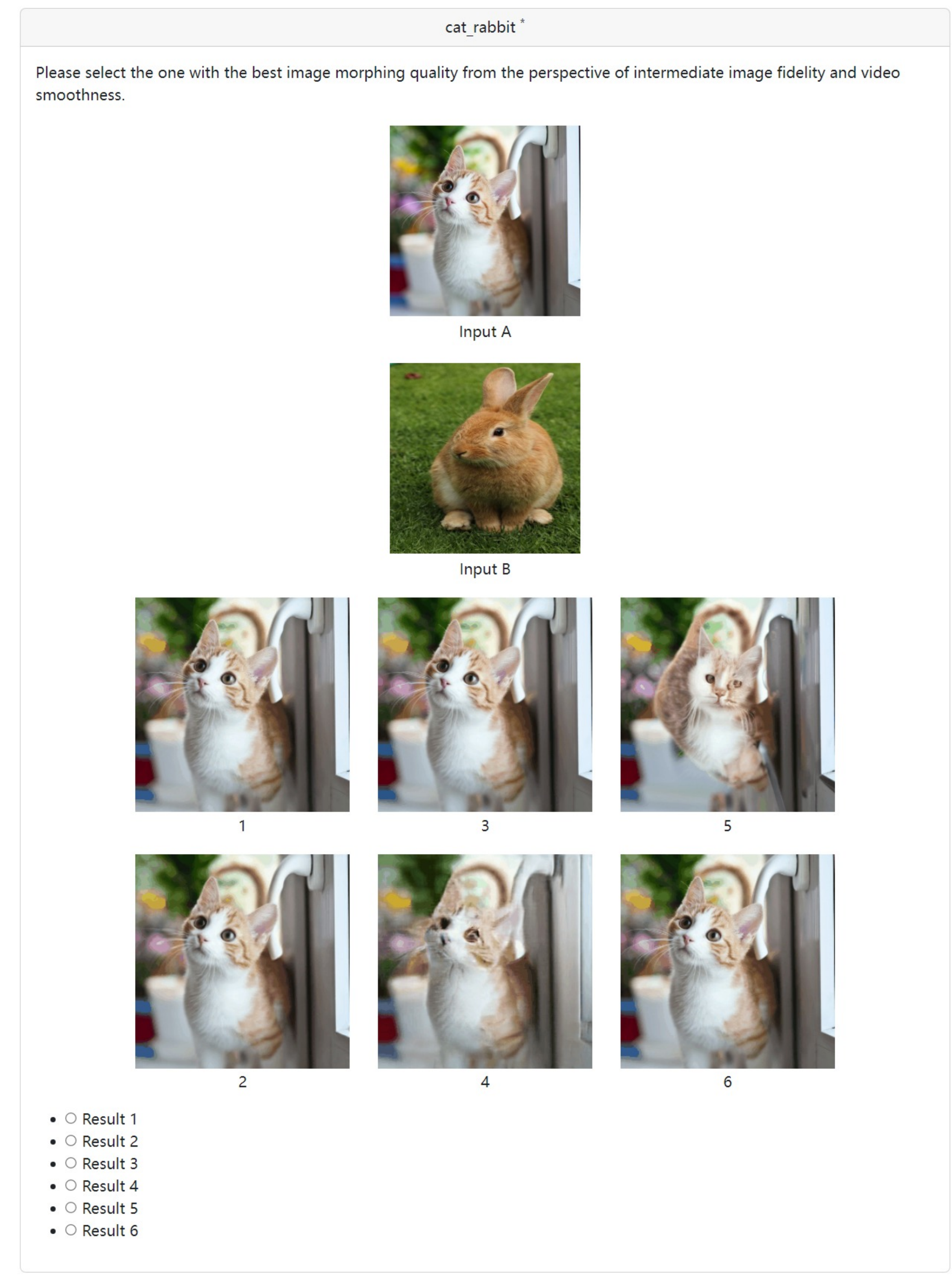}
	\end{center}
	\vspace{-6mm}
	\captionof{figure}{An example of the questionnaire we used in the user study. Note that all the results shown here are videos.}
	\label{fig:question}
	\vspace{6mm}
}]

\end{document}